\documentclass{article}

\usepackage[preprint, nonatbib]{neurips_2024}
\usepackage{array}
\usepackage{longtable}
\usepackage{booktabs}
\usepackage{geometry}

\usepackage[utf8]{inputenc} 
\usepackage[T1]{fontenc}    
\usepackage{hyperref}       
\usepackage{url}            
\usepackage{booktabs}       
\usepackage{amsmath}
\usepackage{amsfonts}       
\usepackage{nicefrac}       
\usepackage{graphicx}
\usepackage{microtype}      
\usepackage{xcolor}         
\usepackage{subcaption}
\usepackage{placeins}
\usepackage{listings}
\usepackage{xcolor}
\usepackage{longtable}

\usepackage[backend=biber,style=authoryear]{biblatex}
\title{Benchmarking the Generality of Vision-Language-Action Models}

\author{%
  \textbf{Pranav Guruprasad  }$^{1,2}$ \quad
  \textbf{Sudipta Chowdhury }$^{1,2}$ \quad
  \textbf{Harsh Sikka }$^{1,2,3}$ \quad
  \textbf{Mridul Sharma }$^{1,7}$ \\[1ex]
  \textbf{Helen Lu }$^{1,4}$ \quad
  \textbf{Sean Rivera }$^{1}$ \quad
  \textbf{Aryan Khurana }$^{1,6}$ \quad
  \textbf{Hangliang Ren }$^{1,5}$ \quad
  \textbf{Yangyue Wang }$^{1,2}$ \\
  \\
  \normalfont
  $^1$Manifold Research \quad $^2$Metarch AI \quad $^3$Georgia Tech \\
  $^4$Tufts University \quad $^5$Northeastern University \\
  $^6$Birla Institute of Technology and Science, Pilani \\
  $^7$Institute for Research and Innovation in Intelligent Systems (IRIIS) \\
  \\
}

\begin{document}

\begin{center}
  \LARGE \bfseries 
  Benchmarking the Generality of Vision-Language-Action Models
  
  \vspace{1.5em} 

  \large 
  \textbf{Pranav Guruprasad}$^{1,2}$, \quad
  \textbf{Sudipta Chowdhury}$^{1,2}$, \quad
  \textbf{Harsh Sikka}$^{1,2,3}$, \\
  \textbf{Mridul Sharma}$^{1,7}$, \quad
  \textbf{Helen Lu}$^{1,4}$, \quad
  \textbf{Sean Rivera}$^{1}$, \quad
  \textbf{Aryan Khurana}$^{1,6}$, \quad
  \textbf{Hangliang Ren}$^{1,5}$, \quad
  \textbf{Yangyue Wang}$^{1,2}$

  \vspace{1.5em} 

  \normalsize \normalfont 
  $^1$Manifold Research, \quad $^2$Metarch AI, \quad $^3$Georgia Tech, \\
  $^4$Tufts University, \quad $^5$Northeastern University, \\
  $^6$Birla Institute of Technology and Science, Pilani, \\
  $^7$Institute for Research and Innovation in Intelligent Systems (IRIIS)
\end{center}

\vspace{2em}

\begin{abstract}
Generalist multimodal agents are expected to unify perception, language, and control—operating robustly across diverse real-world domains. However, current evaluation practices remain fragmented across isolated benchmarks, making it difficult to assess whether today's foundation models truly generalize beyond their training distributions. We introduce MultiNet v1.0, a unified benchmark for measuring the cross-domain generality of vision–language models (VLMs) and vision–language–action models (VLAs) across six foundational capability regimes: visual grounding, spatial reasoning, tool use, physical commonsense, multi-agent coordination, and continuous robot control. Evaluating GPT-5, $\pi_0$, and Magma, we find that no model demonstrates consistent generality: all exhibit substantial degradation on unseen domains, unfamiliar modalities, or cross-domain task shifts—despite strong performance within their training distributions. These failures manifest as modality misalignment, output format instability, and catastrophic knowledge degradation under domain transfer. Our findings reveal a persistent gap between the aspiration of generalist intelligence and the actual capabilities of current foundation models. MultiNet v1.0 provides a standardized evaluation substrate for diagnosing these gaps and guiding the development of future generalist agents. Code, data, and leaderboards are publicly available.
\end{abstract}

\section{Introduction}
Intelligences capable of unifying perception, cognition and control, remains a central challenge in AI. Such systems would be able to perceive multimodal inputs, reason through complex scenarios, and execute actions across diverse real-world domains. Recent advances in large-scale multimodal \parencite{wang2024largescalemultimodalpretrainedmodels} and embodied\parencite{embodied2025} models suggest early progress toward this goal. Models like GPT-4o \parencite{openai2024gpt4ocard} demonstrate remarkable visual and language understanding, specialized models like OpenVLA \parencite{kim24openvla} and $\pi$0 \parencite{black2024pi0visionlanguageactionflowmodel} show impressive robotic manipulation capabilities within their training domains whereas generalist models like Magma \parencite{yang2025magma} perform well on some digital and physical tasks. A fundamental pair of questions remain: "Do these models truly generalize across the diverse modalities and tasks that define real-world intelligence?" Current evaluation practices make it nearly impossible to answer this question with confidence. The landscape of multimodal model assessment remains fragmented, with specialized benchmarks designed for isolated capabilities, i.e., robotics datasets test manipulation in constrained lab settings, vision-language benchmarks focus on static image understanding, and gameplay environments evaluate reactive policies in procedurally generated worlds. Models are optimized and validated within narrow distributions that bear little resemblance to the cross-domain generalization demanded by real-world deployment.

Previous works such as MMMU \parencite{yue2024mmmu}, GAIA \parencite{mialon2023gaia}, BuilderBench \parencite{ghugare2025builderbench}, MultiNet v0.1 \parencite{guruprasad2024benchmarkingvisionlanguage}, and MultiNet v0.2 \parencite{guruprasad2025benchmarkingvisionlanguage} directly exposed this gap. MMMU \parencite{yue2024mmmu} highlighted persistent weaknesses in multimodal understanding across diverse academic tasks requiring cross-domain reasoning. GAIA \parencite{mialon2023gaia} emphasized limitations in real-world task decomposition and tool-augmented problem solving. BuilderBench \parencite{ghugare2025builderbench} revealed that foundation models struggle with spatially grounded, multi-step digital manipulation. MultiNet v0.1 \parencite{guruprasad2024benchmarkingvisionlanguage} exposed severe action misalignment and inconsistent performance across 20 embodied datasets. MultiNet v0.2 \parencite{guruprasad2025benchmarkingvisionlanguage} showed that action-trained VLAs transfer better to discrete environments but fail under out-of-distribution visual and mechanics shifts.
\begin{figure*}
\includegraphics[width=0.98\linewidth]{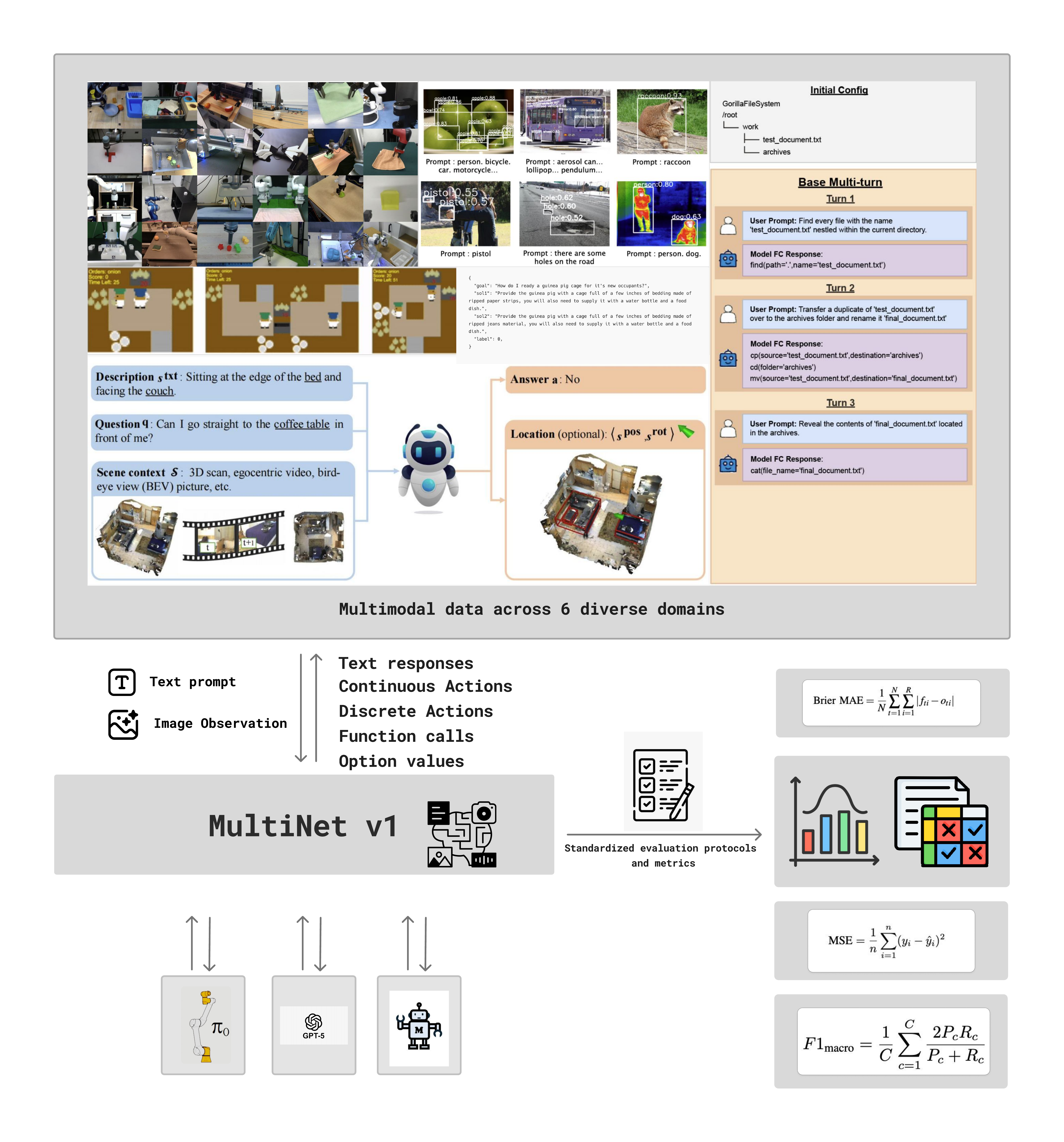}
\caption{Overview of the MultiNet v1.0 benchmark. MultiNet unifies evaluation across core capabilities required for real-world generalist models—including visual grounding, spatial reasoning, physical commonsense reasoning, multi-agent coordination, continuous robot control, and tool-based function calling. The benchmark integrates heterogeneous datasets spanning robotics (Open-X Embodiment), cooperative gameplay (Overcooked), commonsense reasoning (PIQA), 3D spatial QA (SQA3D), in-the-wild grounding (ODINW), and structured API workflows (BFCL). This unified design provides a comprehensive stress test for vision–language and vision–language–action models, exposing representational bottlenecks, modality misalignment, and cross-domain generalization gaps while establishing a standardized substrate for reproducible model comparison and future agent development.}

\label{fig:multinet}
\end{figure*}
Our benchmark was designed to address this gap. 

We identified a core set of foundational capabilities that any generalist model must demonstrate: visual grounding, spatial reasoning, multi-step planning, discrete and continuous action taking, physical commonsense reasoning, function calling, and tool use. These capabilities collectively define the operational space of real-world intelligence. MultiNet v1.0 unifies evaluation across these capabilities systematically: robotics (Open-X-Embodiment \parencite{embodimentcollaboration2025openxembodimentroboticlearning}), cooperative multi-agent gameplay (Overcooked) \parencite{carroll2020utilitylearninghumanshumanai}, physical commonsense reasoning (PIQA) \parencite{bisk2019piqareasoningphysicalcommonsense}, open-world object detection (ODINW) \parencite{xu2023multimodalqueriedobjectdetection}, 3D spatial question answering (SQA3D) \parencite{ma2023sqa3dsituatedquestionanswering}, and conversational function calling (BFCL) \parencite{patil2025bfcl}. We provide an open-source evaluation suite\footnote{\url{https://github.com/ManifoldRG/MultiNet/}}. This unification serves two purposes. First, it provides a stress test for multimodal foundation models, revealing where representational bottlenecks and misalignments occur. Second, it establishes a development substrate for next-generation agents that must operate seamlessly across perception, reasoning, and control. Evaluating them on isolated tasks risks a misleading impression of progress. Only through unified, cross-domain testing can we expose failure modes early and guide model design toward genuine generality.

The initial results from our experiments suggests that there is still a long way to go in order to develop true general models. Pi0 \parencite{black2024pi0visionlanguageactionflowmodel} has completely lost its language generation capability after post-training on action. Magma \parencite{yang2025magma} demonstrates output modality confusion. Even GPT-5 \parencite{gpt52025systemcard}, the strongest general-purpose model, struggle with cooperative multi-agent gameplay, 3D spatial question answering, and conversational function calling.

Our primary contributions in the paper are:
\begin{itemize}
\item Open-source model adaptation code enabling consistent evaluation of diverse architectures across heterogeneous task domains.
\item A standardized submission pipeline that ensures reproducible benchmarking, result validation, and cross-model comparability.
\item Comprehensive results assessing the generality of state-of-the-art multimodal and action-capable models across perception, reasoning, and control tasks.
\item An open-source SDK that streamlines downloading, processing, and translating datasets from a wide range of domains and annotation formats.
\end{itemize}

\section{Related Work}
\subsection{Vision Language Models}

GPT-5 represents the current state-of-the-art in closed-source vision-language models, consolidating strong performance across multiple benchmarks including visual question answering, image understanding, and multimodal reasoning tasks~\parencite{OpenAI2025GPT5SystemCard}. As a large-scale multimodal Transformer trained on diverse internet data, GPT-5 exhibits robust zero-shot capabilities across vision-language tasks. We selected GPT-5 to evaluate whether general-purpose vision-language models trained on web-scale data can effectively handle action-oriented tasks without explicit action training, serving as a critical baseline for understanding the gap between vision-language understanding and action prediction capabilities.

\subsection{Vision Language Action Models}
Vision-language-action models extend multimodal understanding with action prediction for embodied AI. Pi0 employs a flow-matching architecture over pretrained vision-language models, trained on 903 million timesteps comprising approximately 10,000 hours of robotic manipulation data spanning 68 diverse tasks across 7 robot configurations~\parencite{black2024pi0visionlanguageactionflowmodel}. The model predicts continuous actions via flow matching, achieving strong zero-shot generalization on in-domain robotics benchmarks. We selected Pi0 Base as our primary VLA to examine whether robotics-focused training corrupts broader vision-language capabilities.

Our prior work in MultiNet v0.2 evaluated architectural variants including OpenVLA~\parencite{kim2024openvlaopensourcevisionlanguageactionmodel} and $\pi_0$-FAST~\parencite{pertsch2025fastefficientactiontokenization}. OpenVLA is a 7-billion parameter model with a vision encoder that projects visual input into the input space of its LLaMA-2 backbone. The model has been trained on over 907K demonstrations with discrete action tokenization. $\pi_0$-FAST is a $\pi$ variant that uses autoregressive decoding with DCT-based action tokenization. These comparisons reveale how decoding strategies, diffusion sampling versus autoregressive generation, fundamentally shape prediction collapse patterns and overconfidence on out-of-distribution data.

\subsection{Generalist Models}

Magma represents an explicit attempt at building foundation models capable of cross-domain multimodal reasoning and action~\parencite{Yang_2025_CVPR}. Unlike domain-specific VLAs trained exclusively on robotics data, Magma integrates vision, language, and action prediction through multi-task training across diverse domains. We selected Magma to directly test the generalist hypothesis: whether models architecturally designed for cross-domain capabilities can maintain coherent performance across robotics, digital control, and vision-language tasks. Magma's inclusion enables us to identify failure modes specific to generalist architectures, particularly output modality misalignment where models produce inappropriate response formats for given task contexts.

\section{Datasets}

To comprehensively evaluate the multifaceted capabilities of generalist models, our framework leverages a curated suite of benchmarks. These datasets were systematically selected to align with the key capabilities of our evaluation framework, spanning a heterogeneous range of domains—from physical commonsense and visual perception to complex embodied robotics and digital agency. This section details the benchmark chosen for each capability, the rationale for its inclusion, and the systematic process by which considered alternatives were excluded.

\subsection{Physical Agency and Robotic Manipulation}
We selected a curated subset of the Open-X Embodiment collection \parencite{embodimentcollaboration2025openxembodimentroboticlearning}, the largest open-source robotics dataset and a standard pre-training source for leading VLAs including OpenVLA and RT-X. This selection enables in-distribution evaluation for robotics-trained models while testing cross-embodiment generalization, a critical capability for deployable generalist policies. Our subset is curated by selecting the dataset with the most episodes per robot morphology, balancing diversity and size. We considered but excluded DROID \parencite{khazatsky2025droidlargescaleinthewildrobot}, MuJoCo-powered benchmarks \parencite{mujoco2012}, Habitat AI \parencite{savva2019habitatplatformembodiedai}, LIBERO \parencite{liu2023liberobenchmarkingknowledgetransfer}, and CALVIN \parencite{mees2021calvin}. Our exclusion criteria prioritized real-world data, which removed simulators like MuJoCo and Habitat AI. DROID was excluded as its data is largely aggregated within Open-X, and both LIBERO and CALVIN were deemed redundant, as their long-horizon compositional tasks are already represented in our Open-X subset.

\subsection{Gameplay and Simulated Agency}
We selected the Overcooked-AI benchmark \parencite{carroll2020utilitylearninghumanshumanai}, a widely-adopted testbed for cooperative AI and zero-shot coordination research, to specifically evaluate multi-agent coordination, a capability absent from single-agent tasks. Its cooperative gameplay requires planning and adaptation to a partner’s behavior. We also evaluated MineRL \parencite{guss2019minerl}, JARVIS-VLA \parencite{chen2024jarvisvla}, Atari \parencite{bellemare2013atari}, BabyAI \parencite{chevalier2018babyai}, and Procgen \parencite{cobbe2020leveragingproceduralgenerationbenchmark}. The single-agent benchmarks (MineRL, JARVIS-VLA) were not prioritized, as we chose to focus on the unique multi-agent challenge. Atari was excluded due to its non-photorealistic visuals, and BabyAI and Procgen were deemed less challenging than JARVIS-VLA.

\subsection{General World Knowledge and Commonsense Reasoning}
We selected the PIQA benchmark \parencite{bisk2019piqareasoningphysicalcommonsense} to probe intuitive physical commonsense (e.g., material properties, tool use), as its “Goal $\rightarrow$ Solution” format directly simulates action selection. We considered a broad set of alternatives: MMLU \parencite{hendrycks2021measuring}, GPQA \parencite{rein2023gpqa}, Hellaswag \parencite{zellers2019hellaswag}, WinoGrande \parencite{sakaguchi2020winogrande}, the AI2 Reasoning Challenge (ARC) \parencite{clark2018arc}, and VisualComet \parencite{park2020visualcomet}. MMLU and GPQA were excluded as they are text-only and add limited novel value. VisualComet was excluded due to unfiltered content. The remaining benchmarks (Hellaswag, WinoGrande, ARC) were not prioritized as they are text-based and lack PIQA’s specific focus on physical, interactive reasoning.

\subsection{Static Visual Comprehension}
We selected ODinW (Object Detection in the Wild) \parencite{xu2023multimodalqueriedobjectdetection} and adapted it for a zero-shot, in-the-wild detection task. This provides a robust evaluation of fine-grained grounding, a core perceptual skill not yet widely benchmarked for VLAs. The other candidates—MMBench \parencite{liu2023mmbench}, MMMU \parencite{yue2024mmmu}, RefCOCO \parencite{yu2016refcoco}, and CountBench \parencite{kamath2023countbench}—were not selected. MMBench was omitted as it is already widely reported, and CountBench was deemed not as challenging. MMMU and RefCOCO were not aligned with our goal of repurposing a large-scale, in-the-wild detection benchmark, for which ODinW was the ideal choice.

\subsection{Multimodal Association}
We selected SQA3D (Situated Question Answering in 3D) \parencite{ma2023sqa3dsituatedquestionanswering} to test 3D spatial and egocentric reasoning. It uniquely requires a model to interpret its own described position in a 3D scene, a foundational skill for agency. We evaluated numerous 2D alternatives: MSCOCO \parencite{lin2014mscoco}, Flickr30k \parencite{plummer2015flickr30k}, VLUE \parencite{wang2022vlue}, VQA v2 \parencite{goyal2017vqa}, TextVQA \parencite{singh2019textvqa}, GQA \parencite{hudson2019gqa}, RefCOCO \parencite{yu2016refcoco}, Flickr30k Entities \parencite{plummer2015flickr30k}, Surprise3D \parencite{an2024surprise3d}, and MultiBench \parencite{liang2021multibench}. These were ultimately not prioritized. VLUE, for instance, is a benchmark consolidation, and the others (e.g., MSCOCO, VQA v2, RefCOCO) lack the critical 3D, egocentric, and spatial reasoning components that SQA3D provides.

\subsection{Digital Agency and Tool Utilization}
We selected the BFCL v3 benchmark \parencite{patil2025bfcl} (multi-turn base dataset), the standard evaluation for LLM function calling with a widely-referenced public leaderboard, to test robust, multi-step API-calling workflows. Multi-turn function calling tests the model's ability to maintain state and chain dependent operations, capabilities essential for real-world agentic workflows. This was prioritized over OpenFunctions v2 as a more challenging and recent dataset. We also considered API-Bank \parencite{li2023apibank}, Nexus, Tau Bench \parencite{yin2024taubench}, ScreenSpot \parencite{zhang2024screenspot}, AndroidWorld \parencite{rawles2024androidworld}, and OSWorld \parencite{yang2024osworld}. The UI-operation benchmarks (e.g., ScreenSpot) were not prioritized, and BFCL v4 was not yet available. We chose to focus on programmatic, multi-turn API calling, making BFCL v3 the most suitable choice.

\subsection{Spatiotemporal Reasoning and Video Understanding}
We selected RoboVQA \parencite{sermanet2023robovqa} for its unique combination of video-conditioned reasoning, real-world grounding, and long-horizon task diversity. The dataset contains 829,502 (video, text) pairs spanning 29,520 unique instructions collected across multiple embodiments (robot, human, human with grasping tool) in real office environments, enabling evaluation of cross-embodiment transfer and temporal reasoning over action sequences. We considered Ego4D \parencite{grauman2022ego4d}, EpicKitchens \parencite{damen2018scaling}, ActivityNet-QA \parencite{yu2019activitynet}, and TEACh \parencite{padmakumar2022teach}. Ego4D and EpicKitchens were excluded as they lack robot embodiment data and focus on human activities rather than robot-executable tasks. TEACh was excluded because it uses simulated environments rather than real-world data. RoboVQA's diverse task types, including planning, success classification, and affordance detection, directly probe capabilities essential for embodied agents operating over extended time horizons.

\section{Methodology}

This section details the evaluation design of MultiNet v1.0, including the adopted metrics, model-adaptation pipeline, and prompting configuration. Following the conventions of Multinetv0.1, all evaluations were performed in a zero-shot setting \parencite{guruprasad2024benchmarkingvisionlanguage}.

\subsection{Metrics}
MultiNet evaluates VLA and VLM models across diverse modalities, each requiring domain-specific metrics. We employ three evaluation suites: discrete action prediction for game environments (Overcooked), continuous action prediction for robotics (OpenX), and vision-language understanding for reasoning and detection tasks (e.g., PIQA, ODINW, RoboVQA, SQA3D). Discrete tasks use categorical actions, continuous tasks use real-valued vectors with varying scales, and vision-language tasks generate text requiring parsing and validation. For each domain, we report primary accuracy metrics alongside error distribution statistics and invalid prediction rates.

For discrete action prediction in game environments (Procgen, Overcooked), near-misses receive no partial credit in discrete control. We employ Exact Match Rate (EMR) as the primary metric, measuring the percentage of predictions that exactly match ground truth actions:
\begin{equation}
\text{EMR} = \frac{\text{Number of correct predictions}}{N}
\end{equation}
For discrete action spaces where each action produces fundamentally different outcomes, EMR provides a strict measure of prediction correctness across Procgen, Overcooked, PIQA multiple-choice questions, and classification-formulated vision tasks. High EMR indicates the model accurately predicts the correct action, while low EMR suggests the model frequently selects incorrect actions regardless of how semantically similar they might be to the ground truth.

We supplement EMR with micro-averaged precision, recall, and F1 scores, which aggregate metrics treating each prediction equally and reflecting overall prediction accuracy weighted by action frequency:
\begin{equation}
P_{\text{micro}} = \frac{\sum_{c} \text{TP}_c}{\sum_{c} (\text{TP}_c + \text{FP}_c)}, \quad R_{\text{micro}} = \frac{\sum_{c} \text{TP}_c}{\sum_{c} (\text{TP}_c + \text{FN}_c)}, \quad F1_{\text{micro}} = \frac{2 P_{\text{micro}} R_{\text{micro}}}{P_{\text{micro}} + R_{\text{micro}}}
\end{equation}
For multi-class single-label prediction tasks where action classes are mutually exclusive, an incorrectly predicted action is both a false positive for its predicted class and a false negative for the ground truth class. Micro metrics aggregate true positives, false positives, and false negatives globally across all timesteps before computing precision, recall, and F1. High micro precision indicates the model gets a large proportion of predictions correct overall, while low micro precision suggests frequent mismatches with ground truth. Micro metrics favor majority classes due to frequency weighting.

Macro-averaged precision, recall, and F1 compute per-class metrics first, then average across all classes, weighting each action class equally regardless of frequency:
\begin{equation}
P_{\text{macro}} = \frac{1}{C}\sum_{c=1}^{C} \frac{TP_c}{TP_c + FP_c}, \quad R_{\text{macro}} = \frac{1}{C}\sum_{c=1}^{C} \frac{TP_c}{TP_c + FN_c}, \quad F1_{\text{macro}} = \frac{1}{C}\sum_{c=1}^{C} \frac{2P_cR_c}{P_c + R_c}
\end{equation}
This reveals systematic biases toward specific actions or failure on rare but critical actions. We primarily report macro recall, which avoids both the majority-class bias of micro metrics and the rare-class sensitivity of macro precision or F1 \parencite{guruprasad2025benchmarkingvisionlanguage}. Macro metrics are particularly valuable for detecting whether models handle rare but strategically important actions (e.g., specialized game actions, object interactions) as reliably as common actions (e.g., basic movement). Low macro recall combined with high micro recall indicates the model performs well on frequent actions but poorly on rare ones. Large discrepancies between micro and macro metrics indicate aggregate performance masking poor rare-action understanding.

We additionally report invalid predictions (outputs outside the valid action space) and calculate precision excluding these, separating formatting from semantic errors \parencite{guruprasad2025benchmarkingvisionlanguage}. Invalid predictions occur when models generate outputs that do not correspond to any valid action in the action space. For VLMs producing probability distributions, outputs are invalid if they fail to parse into the expected format (e.g., malformed JSON, probabilities not summing to 1). A high invalid percentage indicates the model struggles to constrain outputs to the target action space, representing a distinct failure mode from semantic misclassification. Precision calculated excluding invalid predictions reveals model performance when outputs are well-formed, isolating semantic understanding from formatting issues. For Overcooked's multi-agent coordination, we decompose the joint action space into per-player accuracies, diagnosing whether coordination failures stem from individual behavior errors or misunderstanding inter-agent dependencies. High individual player accuracy combined with low joint accuracy suggests the model understands each agent's behavior independently but fails to capture coordination patterns.

For continuous action prediction in robotics tasks from OpenX, raw error values are incomparable across datasets due to heterogeneous action spaces with different scales. We employ Mean Absolute Error (MAE) and Mean Squared Error (MSE) as base metrics, computing per-timestep distance averaged across action dimensions:
\begin{equation}
\text{MAE} = \frac{1}{N} \sum_{i=1}^{N} |\hat{\mathbf{a}}_i - \mathbf{a}_i|, \quad \text{MSE} = \frac{1}{N} \sum_{i=1}^{N} \|\hat{\mathbf{a}}_i - \mathbf{a}_i\|^2
\end{equation}
MAE provides interpretable error magnitude in the action space's native units, while MSE penalizes large deviations more heavily by squaring errors, making it sensitive to outliers. Low MAE indicates predicted actions are close to ground truth on average, while high MAE suggests significant deviations. However, raw MAE and MSE values are not comparable across datasets with different action space scales (e.g., joint angles in radians vs. gripper forces in Newtons).

We therefore normalize model errors relative to a baseline predictor that always outputs the training set mean action for each dimension, yielding Approximate Relative MAE (Approx RelMAE) as our primary cross-dataset comparison metric:

\begin{equation}
\text{Approx RelMAE} = \frac{\frac{1}{N} \sum{j=1}^{N} \text{MAE}{\text{model}, j}}{\frac{1}{N} \sum{j=1}^{N} \text{MAE}{\text{baseline}, j}} = \frac{\sum{j=1}^{N} \sum{i=1}^{D} | \hat{y}{j,i} - y{j,i} |}{\sum{j=1}^{N} \sum{i=1}^{D} | \bar{y}{\text{train}, i} - y{j,i} |}
\end{equation} where $\hat{y}_{j,i}$ is the model's predicted action for dimension $i$ at timestep $j$, $y_{j,i}$ is the ground truth action, $\bar{y}_{\text{train}, i}$ is the mean action value for dimension $i$ computed over the training set, $N$ is the number of timesteps, and $D$ is the action dimensionality. The baseline predictor represents an uninformed strategy that always predicts the central tendency of the training distribution regardless of visual input or context. Approx RelMAE values less than 1.0 indicate the model outperforms this naive baseline, while values greater than 1.0 indicate worse-than-baseline performance. This normalization enables cross-dataset comparison by measuring model performance relative to dataset-specific baselines rather than using absolute error scales. For invalid predictions (NaN, inf, or malformed outputs), we assign the baseline error for that timestep, ensuring invalid outputs receive appropriate penalties without arbitrarily inflating metrics.

Continuous control errors exhibit fat-tailed distributions with occasional catastrophic predictions, so we additionally report quantile-filtered Approx RelMAE (excluding errors beyond the 5th and 95th percentiles), maximum relative MAE, and the proportion of predictions exceeding 3x the median error threshold. These outlier-specific metrics capture failure modes that mean-based metrics obscure. Quantile-filtered Approx RelMAE reveals typical performance when extreme outliers are excluded. Maximum relative MAE quantifies how much the worst-case error deviates from the median error; values significantly greater than 1 indicate some extremely poor predictions that skew mean metrics. The proportion of predictions exceeding 3x median threshold identifies the frequency of severe failures.

\begin{equation}
 \text{baseline} = \text{mean}(|\mathbf{a}_{\max} - \mathbf{a}_{\min}|)   
\end{equation}
This represents the worst-case uninformed error, equivalent to always predicting the maximum action value when the ground truth is minimum (or vice versa). By including these baseline penalties in the normalization distribution, NAMAE measures model performance relative to this worst-case baseline, where 0.0 represents perfect predictions and 1.0 represents baseline-level errors.

Continuous control errors exhibit fat-tailed distributions with occasional catastrophic predictions, so we additionally report quantile-filtered NAMAE (excluding errors beyond the 5th and 95th percentiles), maximum relative MAE, and the proportion of predictions exceeding 3x the median error threshold. These outlier-specific metrics capture failure modes that mean-based metrics obscure. Quantile-filtered NAMAE reveals typical performance when extreme outliers are excluded. Maximum relative MAE quantifies how much the worst-case error deviates from the median error; values significantly greater than 1 indicate some extremely poor predictions that skew mean metrics. The proportion of predictions exceeding 3x median threshold identifies the frequency of severe failures.

Continuous control errors exhibit fat-tailed distributions with occasional catastrophic predictions, so we additionally report quantile-filtered NAMAE (excluding errors beyond the 5th and 95th percentiles), maximum relative MAE, and the proportion of predictions exceeding 3x the median error threshold. These outlier-specific metrics capture failure modes that mean-based metrics obscure. Quantile-filtered NAMAE reveals typical performance when extreme outliers are excluded. Maximum relative MAE quantifies how much the worst-case error deviates from the median error; values significantly greater than 1 indicate some extremely poor predictions that skew mean metrics. The proportion of predictions exceeding 3x median threshold identifies the frequency of severe failures.

For vision-language tasks, we employ task-specific metrics based on output format. Exact Match Rate (EMR) measures the percentage of successfully parsed predictions that exactly match ground truth answers for tasks with discrete answer sets. Tasks span diverse formats: multiple-choice selection (PIQA), free-form text responses (SQA3D, RoboVQA), structured function calls (BFCL), and object classification. For ODINW, we reformulate the object detection task as a classification problem where models predict object category presence rather than bounding box coordinates, enabling unified evaluation across VLMs and VLAs. We parse outputs into expected formats, flagging parsing failures or constraint violations as invalid. High EMR indicates strong task understanding and correct output generation, while low EMR suggests either semantic misunderstanding or formatting errors. We distinguish between these failure modes by separately tracking invalid predictions.

For free-form vision-language tasks where multiple valid textual responses exist, we measure semantic similarity using cosine similarity between sentence embeddings of predicted and reference answers:
\begin{equation}
\text{Sim}(\hat{y}, y) = \frac{\text{emb}(\hat{y}) \cdot \text{emb}(y)}{\|\text{emb}(\hat{y})\| \|\text{emb}(y)\|}
\end{equation}
Embeddings are computed using all-MiniLM-L6-v2. This metric captures semantic similarity even when predicted answers use different wording than reference answers for tasks like SQA3D and RoboVQA. Similarity scores near 1.0 indicate high semantic overlap, while scores near 0 indicate unrelated or contradictory answers. This provides a more nuanced evaluation than strict string matching for tasks where multiple phrasings are valid. We report invalid prediction rates separately, as formatting failures represent a distinct failure mode from semantic errors. Models may demonstrate correct reasoning but fail to produce outputs in the required structure, or conversely, may generate well-formed but semantically incorrect outputs.

We employ diagnostic tools to isolate failure modes. For discrete action tasks, we report clipped metrics where invalid predictions are clipped to valid action space bounds, establishing performance upper bounds via output normalization. Large gaps between clipped and unclipped metrics indicate that output formatting issues, rather than semantic understanding, limit performance. For probabilistic predictions in multi-agent tasks like Overcooked, we quantify calibration using Brier MAE, which measures mean absolute error between predicted probability distributions and one-hot ground truth:
\begin{equation}
\text{Brier MAE} = \frac{1}{N} \sum_{t=1}^{N} \sum_{i=1}^{R} |p_{ti} - y_{ti}|
\end{equation}
where $p_{ti}$ is the predicted probability for class $i$ at timestep $t$, $y_{ti}$ is the one-hot ground truth, and $R$ is the number of classes. Following \parencite{guruprasad2025benchmarkingvisionlanguage}, we use MAE rather than MSE to maintain interpretability while reducing outlier sensitivity. This metric penalizes confident but incorrect predictions more severely than uncertain errors, assessing whether models assign high probability to likely actions and low probability to unlikely alternatives based on context. Good Brier scores combined with poor EMR indicate well-calibrated uncertainty but indecisive predictions, while poor Brier scores with good EMR suggest overconfident but occasionally correct predictions. Brier MAE has a maximum value of 2, achieved when the model assigns probability 1 to an incorrect action.

{
\footnotesize 
\setlength{\tabcolsep}{4pt} 
\renewcommand{\arraystretch}{1.3} 

\begin{longtable}{
    >{\raggedright\arraybackslash}p{0.13\textwidth}  
    >{\raggedright\arraybackslash}p{0.20\textwidth}  
    >{\raggedright\arraybackslash}p{0.23\textwidth}  
    >{\raggedright\arraybackslash}p{0.12\textwidth}  
    >{\raggedright\arraybackslash}p{0.14\textwidth}  
    >{\raggedright\arraybackslash}p{0.10\textwidth}  
}
\caption{Prompt templates and expected outputs used in the MultiNet v1.0 benchmark. Example outputs are representative of correctly formatted responses. Prompt templates are slightly different among models.}
\label{tab:prompt_templates}\\
\toprule
\textbf{Dataset} & \textbf{Prompt Template} & \textbf{Ex. Input} & \textbf{Out. Format} & \textbf{Ex. Output} & \textbf{Metric} \\
\midrule
\endfirsthead

\multicolumn{6}{c}{{\bfseries \tablename\ \thetable{} -- continued from previous page}} \\
\toprule
\textbf{Dataset} & \textbf{Prompt Template} & \textbf{Ex. Input} & \textbf{Out. Format} & \textbf{Ex. Output} & \textbf{Metric} \\
\midrule
\endhead

\hline
\multicolumn{6}{r}{{Continued on next page}} \\
\endfoot

\bottomrule
\endlastfoot

\textbf{PIQA} & 
You are evaluating physical commonsense. Given two choices, select the physically plausible one. Output only the index of the correct solution, and nothing else...\textit{<Question>} &
\textit{Question:} "goal": "How to make a simple ice pack at home", "sol1": "Take a clean sock and fill it with uncooked rice. Tie the end and place it in the freezer for a few hours.", "sol2": "Take a clean sock and fill it with cooked pasta. Tie the end and place it in the freezer for a few hours.", &
Binary choice (\texttt{0} or \texttt{1}) &
\texttt{0} &
EMR
\\
\midrule

\textbf{SQA3D} & 
You are a vision-language model specializing in 3D scene understanding and question answering... Instructions: Respond with only your answer. Do not provide explanations or reasoning \textit{<Scene>}\textit{<Question>} &
"situation": "I am standing by the ottoman on my right facing a couple of toolboxes.", "alternative\_situation": "I just placed two backpacks on the ottoman on my right side before I went to play the piano in front of me to the right.", "question": "What instrument in front of me is ebony and ivory?", "image": "<scene image as numpy array with shape (H, W, 3)>" &
Short natural-language phrase &
\texttt{"piano"} &
EMR \& Semantic Similarity
\\
\midrule

\textbf{RoboVQA} & 
You are a Visual-Language Model Assistant that specializes in answering questions about robotics tasks based on given images representing the robot's environment... Output formatting rules:... \textit{<Image>}\textit{<Question>} &
"question": "Task and Context: Navigate to the kitchen counter and pick up the red mug. Question: What should I do next?", "image": "<robot's camera view as numpy array with shape (H, W, 3)>" &
Yes/No/ Action phrase &
\texttt{"grasp the red mug handle"} &
EMR \& Semantic Similarity
\\
\midrule

\textbf{ODinW} & 
You are a specialized Visual-Language Model Assistant that identifies the object in a given image and selects the best option possible from the options provided... The answer must be a single integer... \textit{<Question>} &
\textit{Question:} "image": "<cropped object image as numpy array with shape (H, W, 3)>", "question": "What object is shown in this image from the BCCD dataset? Option 0: RBC Option 1: WBC Option 2: Platelets Output the number (0-2) of the correct option only.", "category\_name": "RBC", "options": ["RBC", "WBC", "Platelets"], &
Single integer label (\texttt{0, 1, 2...}) &
\texttt{"0"} &
EMR \& F1
\\
\midrule

\textbf{BFCL} & 
You are an AI assistant that can call functions to complete tasks. You will be presented with conversation histories where each turn may require function calls... Output only the function calls, no explanations or additional text. \textit{<Conversation>} &
\textbf{check the Appendix for an example conversation A\ref{lst:bfcl_multiturn}} &
List of structured function calls &
\texttt{[mkdir \newline ("data"), \newline move("a.txt" \newline , "data/")]} &
EMR \& Semantic Similarity
\\
\midrule

\textbf{Overcooked (Action)} & 
Layout \textit{<Layout>}. Time: \textit{<time>} elapsed... Actions: 0-5: Player 0: North... Output joint action index as single value between 0-35. &
\textit{Observation:} "layout\_name": "cramped\_room", "image\_observation": "<game screenshot as numpy array with shape (H, W, 3)>", "time\_elapsed": 19.5, &
Single integer label (\texttt{0, 1...}) &
\texttt{0} &
EMR \& F1
\\
\midrule

\textbf{Overcooked (Probs)} & 
We are running a simulation for two AI agents cooperatively playing Overcooked in layout \textit{<Layout>}, a kitchen coordination game. \textit{<Observation>} &
- &
List of 36 probs &
\texttt{[0.12, 0.04, ..., 0.03]} &
BNAMAE
\\
\midrule

\textbf{Open-X Embodiment} & 
You are an AI agent performing the task \textit{<Input>}. Predict the next action in the control sequence. &
"text\_observation": "pick up the red block and place it in the bin", "image\_observation": "<robot camera view as numpy array with shape (H, W, 3)>" &
Continuous 7D or 8D action vector &
\texttt{[0.12, -0.05, 0.08, 0.0, 0.0, 0.3, 1.0]} &
BNAMAE \& BNAMSE
\\
\midrule

\textbf{Open-X Quadrupedal} & 
You are an AI agent to solve the task... The actions available: \textit{<Action Schema with Min/Max/Mean Stats>}... You must generate your output keeping the following format: A list starting with '['...'' &
- &
Continuous 7D or 8D action vector &
\texttt{[0.15, -0.07, 0.02, ..., 1.0]} &
BNAMAE \& BNAMSE
\\

\end{longtable}
}

\subsection{Model Adaptation}
\subsubsection{$\pi_0$ Adaptation for Language Generation}
The $\pi_{0}$ model is a large-scale vision-language-action (VLA) system that builds on the PaliGemma-3B \parencite{beyer2024paligemmaversatile3bvlm} vision-language backbone but replaces its autoregressive language-generation loop with a downstream action expert head to enable continuous robot control. Its architecture consists of:

\begin{itemize}
    \item \textbf{Vision Encoder:} A SigLIP-based image tower that converts visual patches into a unified multimodal token sequence.
    \item \textbf{Language Backbone:} A Gemma-style transformer decoder that fuses visual and textual embeddings through cross-attention, producing multimodal latent representations.
    \item \textbf{Action Expert:} A continuous-output projection head replacing the original text-generation head, mapping the final hidden states to parameterized action vectors (\textit{e.g.}, end-effector position, orientation, and gripper control).
\end{itemize}

Unlike standard vision-language models, $\pi_{0}$ was structurally adapted for robotic action prediction. Specifically, the autoregressive decoding and token-generation layers are disconnected from the vision-language backbone. This modification renders the model incapable of producing natural-language outputs.

To evaluate $\pi_{0}$’s underlying vision-language representations on language tasks, we restored text-generation capability through a simple weight-injection procedure. Rather than modifying the model architecture, we directly replaced the internal weights of a Hugging Face paligemma-3b-pt-224 checkpoint--the variant used as the backbone of $\pi_0$--with those from $\pi_{0}$’s PaliGemma backbone. The process proceeds as follows:

\begin{enumerate}
    \item Load the Hugging Face paligemma-3b-pt-224 model, which includes the complete text-generation loop (tokenizer, autoregressive decoding, and language projection head).
    \item Extract the vision-language backbone parameters from the adapted $\pi_{0}$ model.
    \item Replace the corresponding backbone weights in the Hugging Face PaliGemma-3B-PT-224 model with those from $\pi_{0}$, leaving the Hugging Face language head and generation loop unchanged.
\end{enumerate}

This produces a model that is architecturally identical to the original Hugging Face PaliGemma-3B-PT-224, but whose backbone weights originate from $\pi_{0}$. The resulting model has $\pi_{0}$’s backbone weights while leveraging Hugging Face’s existing autoregressive generation module, allowing us to extract text tokens from $\pi_0$'s paligemma backbone.

\subsubsection{GPT-5 Prompt Adaptation for Action Outputs}
To bridge the gap between GPT-5's general-purpose text generation and the strict requirements of embodied control, we used a prompting strategy that aligns the model's outputs with the specific action spaces of our benchmarks.

\paragraph{Overcooked (Discrete Joint Actions)}
In the multi-agent Overcooked environment, we adapted the prompt to produce a full probability distribution over the joint action space rather than a single discrete action label. The model was explicitly instructed to generate a list of exactly 36 decimal values, corresponding to the probability of each possible joint action tuple (e.g., \texttt{Player 0: NORTH, Player 1: STAY}). To ensure validity, the system prompt enforced a strict constraint that these values must sum exactly to 1.0. This adaptation allowed us to evaluate the calibration and uncertainty of the model's cooperative planning using Brier scores, providing a more granular assessment than simple exact-match accuracy.

\paragraph{Open-X Embodiment (Continuous Control)}
For robotic manipulation tasks in the Open-X dataset, where the output is a continuous vector, we employed a dynamic prompting strategy to ground the model's numerical predictions. The system prompt was constructed by parsing the dataset definitions:
\begin{itemize}
    \item \textbf{Verbal Descriptions:} For dimensions with clear semantic meaning (e.g., "Gripper open or closed"), the prompt explicitly mapped values to descriptions (e.g., "1.0: Gripper closed").
    \item \textbf{Statistical Grounding:} For abstract continuous dimensions without verbal labels, the prompt provided the statistical distribution of the action space derived from the training set. This included the minimum, maximum, and mean values for each dimension (e.g., "Range: -0.05 $\sim$ 0.05, Mean: 0.002").
\end{itemize}
This context served to scale the model's outputs within physically executable limits, lightening the difficulty of zero-shot continuous value prediction.

\subsubsection{Magma Adaptation for Action Output Dimensions}
Magma's architecture includes a specialized action head trained primarily for 7-degree-of-freedom (7-DoF) robotic manipulation tasks (typically end-effector pose plus gripper). This architectural specialization prevents direct zero-shot transfer to environments with incompatible action spaces, such as the 12-dimensional joint control required for the \texttt{utokyo\_saytap} quadrupedal robot or the discrete coordination space of Overcooked.

To evaluate Magma on these out-of-distribution action spaces, we bypassed the specialized action head and utilized the model's underlying vision-language capabilities to predict actions as text.

\paragraph{Overcooked}
Similar to the GPT-5 adaptation, the model was instructed to analyze the game state image and textually generate a string representation of a list containing 36 probability values. These text outputs were then parsed and validated against the unit-sum constraint to compute the Brier score metrics.

\paragraph{High-DoF Robotics (Open-X Quadrupedal)}
For the 12-dimensional control tasks in the \texttt{utokyo\_saytap} dataset, we treated the control task as a text generation problem. The system prompt was augmented with a semantic map of the 12-dimensional action vector, explicitly identifying each index (e.g., "Motor joint position for front right hip", "rear left calf"). In addition, to guide the model's generation of raw motor positions, we included the valid numerical range (min/max) and mean values for each joint derived from the dataset statistics. Magma was instructed to output a text-based list of 12 floating-point numbers, which, if successful, could be parsed into the continuous action vector for evaluation.


\section{Results and Discussion}

\begin{figure*}[]
\centering
\includegraphics[width=0.98\linewidth]{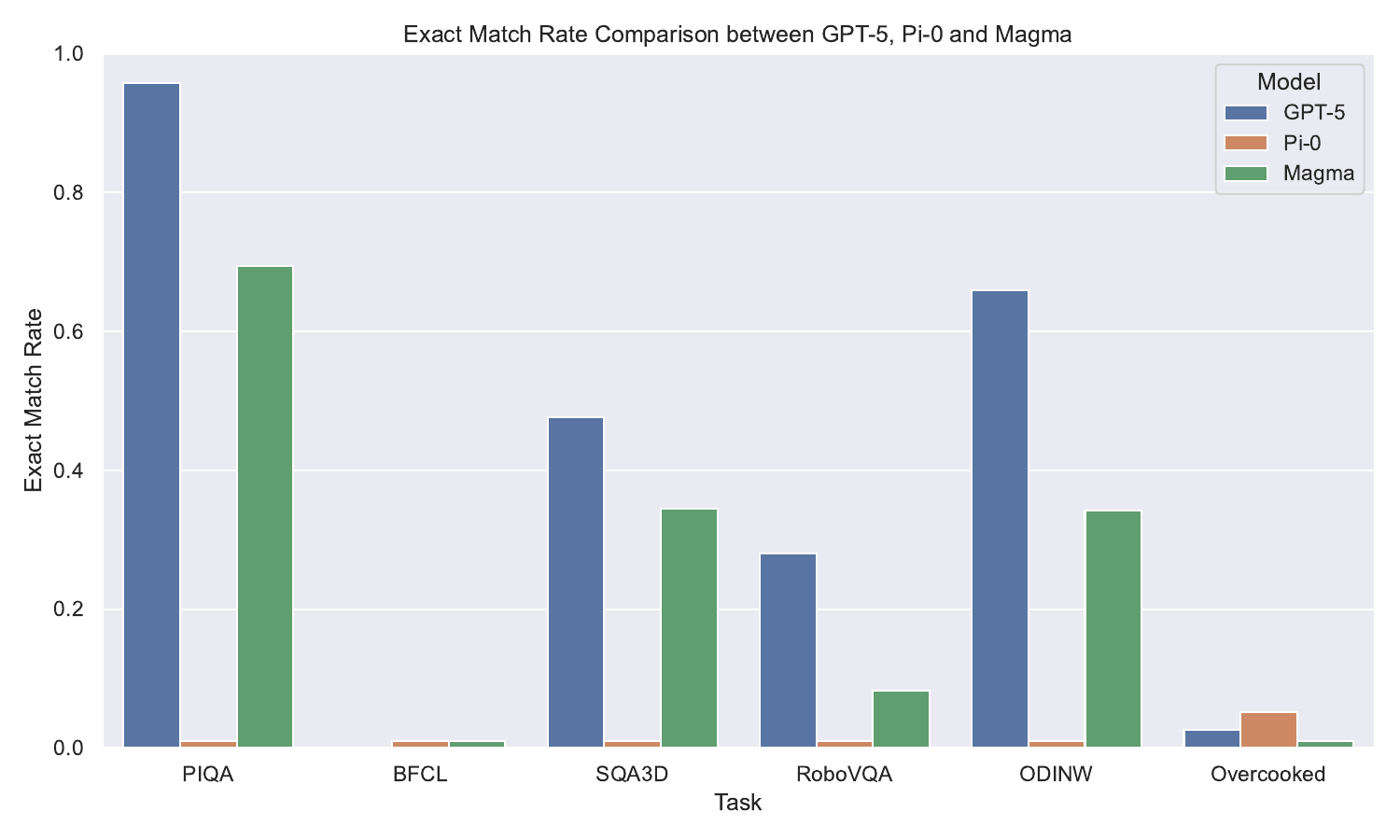}
\caption{Exact Match Rate (EMR) comparison across all vision–language tasks (PIQA, SQA3D, RoboVQA, ODINW, BFCL). GPT-5 achieves the highest EMR across datasets, Magma shows moderate performance with frequent output-format errors, and $\pi_{0}$ collapses on all VL tasks due to losing language generation capability. GPT-5’s BFCL score is taken from the leaderboard (37.38\%).}
\label{fig:emr}
\end{figure*}

\begin{figure*}[]
\centering
\includegraphics[width=0.98\linewidth]{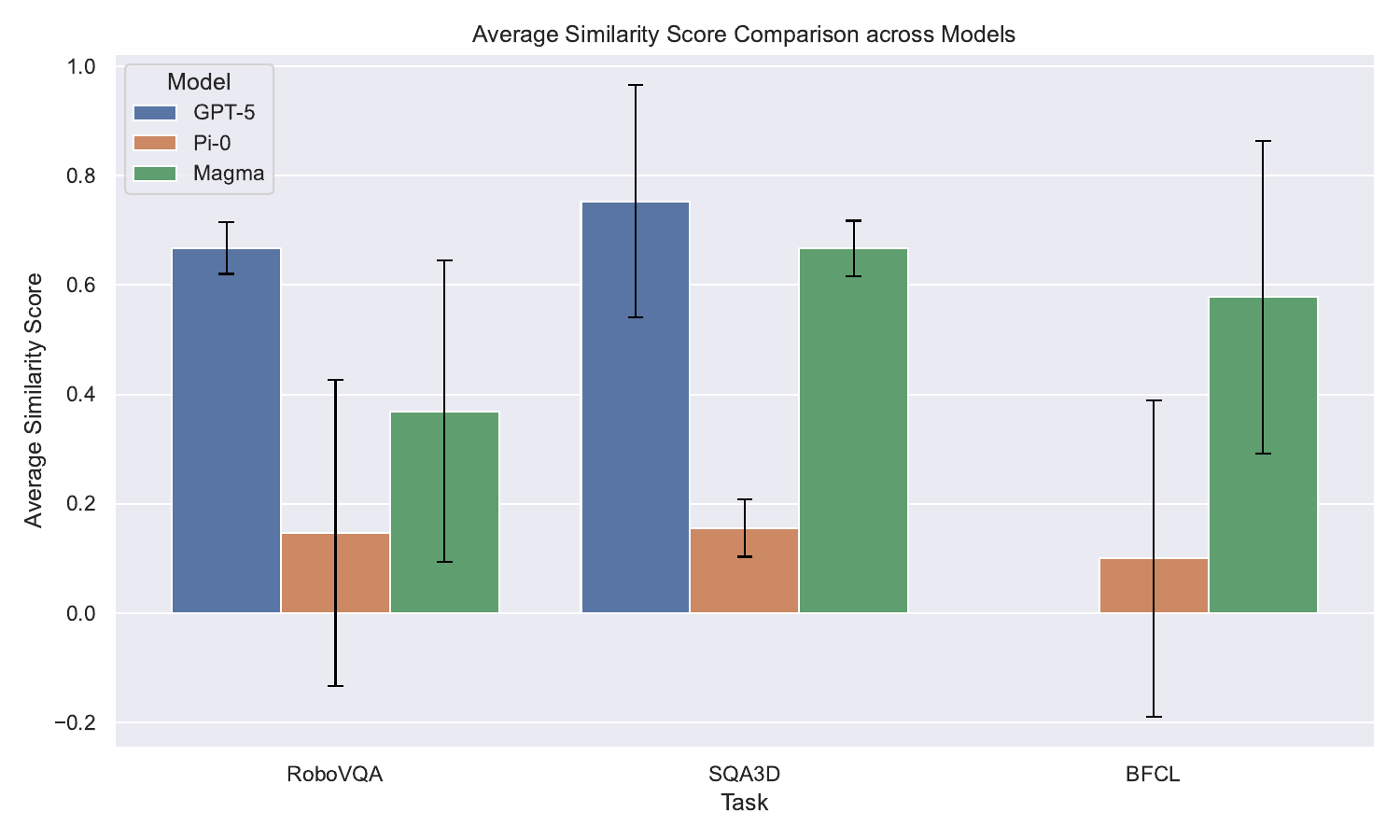}
\caption{Average semantic-similarity score across VL tasks. GPT-5 maintains the most stable semantic alignment; Magma is moderate; $\pi_{0}$ collapses. Note that we did not profile GPT-5 on BFCL as its BFCL performance can be referenced from the BFCL leaderboard \parencite{patil2025bfcl}}
\label{fig:similarity}
\end{figure*}

\begin{figure*}[]
\centering
\includegraphics[width=0.98\linewidth]{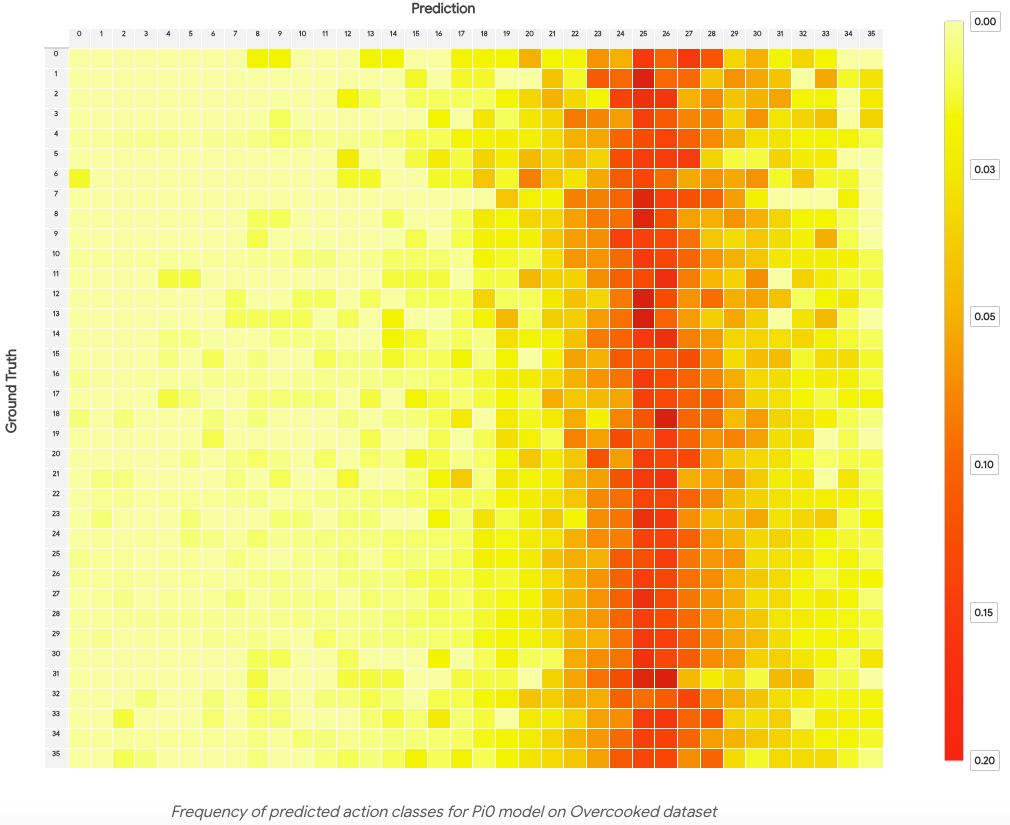}
\caption{Pi0 experiences prediction space collapse on the Overcooked dataset, centered around the actions 25 and 26, which maps to (Player 1: STAY, Player 2: SOUTH) and (Player 1: STAY, Player 2: EAST) respectively.}
\label{fig:pi0_action_collapse}
\end{figure*}

We evaluate three representative models—GPT-5 (general VLM), $\pi_0$ (robotics-tuned VLA), and Magma (generalist multimodal agent)—across vision–language (VL), discrete multi-agent control, and continuous robot control. Unless noted, evaluations are zero-shot with task-specific prompting and strict output parsing.

\subsection{Vision–Language Accuracy (EMR and Semantic Similarity)}

As shown in Fig.~\ref{fig:emr}, GPT-5 attains the highest exact-match rate (EMR) across PIQA, SQA3D, RoboVQA, ODINW, and BFCL, reflecting strong language grounding and format compliance. Note that we did not profile GPT-5 on BFCL. Instead, we reference its result from the BFCL leaderboard \parencite{patil2025bfcl}. According to the leaderboard, the GPT-5-2025-08-07 (Prompt) variant achieves an overall accuracy of 37.38\% with multi-turn prompting. In contrast, $\pi_0$ collapses on VL outputs: EMR is near zero across tasks, consistent with catastrophic loss of general language generation ability after post-training on action. Specifically, we find that $\pi_0$ tends to repeat the "increa" token on vision language tasks. Magma sits between GPT-5 and $\pi_0$ on EMR but often emits the wrong type of output i.e. coordinates instead of labels), depressing exact-match despite partially relevant content. 

While EMR focuses on strict format, semantic similarity provides a more forgiving measure for tasks with natural language answers (SQA3D, RoboVQA, and BFCL). GPT-5 maintains its strong performance on this metric, confirming the semantic accuracy of its outputs. Similarly, Magma sits between GPT-5 and $\pi_0$ in terms of semantic similarity, supporting the finding that it often produces semantically relevant though sometimes incorrectly formatted content i.e. coordinates instead of labels. The $\pi_0$ model, however, shows near-zero semantic similarity scores, confirming the catastrophic collapse of its language generation ability.

\subsection{Object Detection and Multi-agent Cooperative Gameplay Discrete Action as Classification (F1)}

\begin{figure*}[]
\centering
\includegraphics[width=0.98\linewidth]{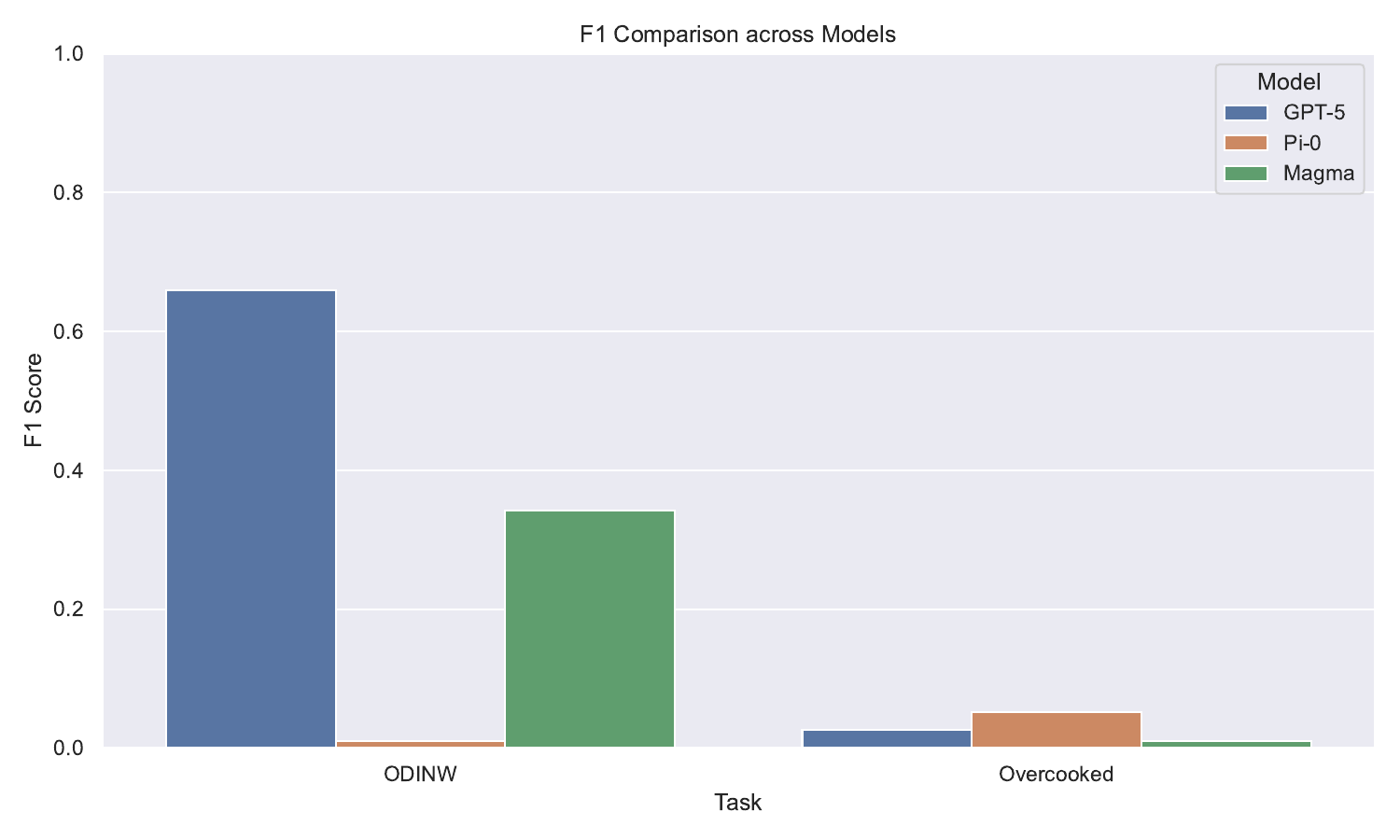}
\caption{F1 comparison for ODINW and Overcooked. GPT-5 maintains superior precision–recall balance, while Magma outputs coordinate strings and $\pi_{0}$ produces gibberish.}
\label{fig:f1}
\end{figure*}

Figure~\ref{fig:f1} reports macro F1 scores for ODINW and Overcooked. GPT-5 achieves the highest F1 on ODINW. However, all three models display low F1 scores on Overcooked. Magma intermittently emits set-of-mark spatial coordinate strings instead of discrete labels \parencite{yang2025magma}. These modality-misaligned errors result in low macro F1.

In addition to low F1 performance, we observe that $\pi_0$ undergoes a pronounced action collapse failure mode during Overcooked evaluation. Instead of producing discrete action labels, $\pi_0$ repeatedly emits a narrow range of actions irrespective of input observation. Specifically, action 25 and action 26 are the most frequently produced action outputs regardless of ground truth. This collapse is visualized in Fig.~\ref{fig:pi0_action_collapse}. This behavior indicates that the model’s action head loses the ability to meaningfully condition on state information. This collapse is consistent with $\pi_0$'s uniformly low macro-F1 scores across discrete control tasks.

\subsection{Continuous Control on Open-X (BNAMAE and BNAMSE)}

\begin{figure*}[]
\centering
\includegraphics[width=0.98\linewidth]{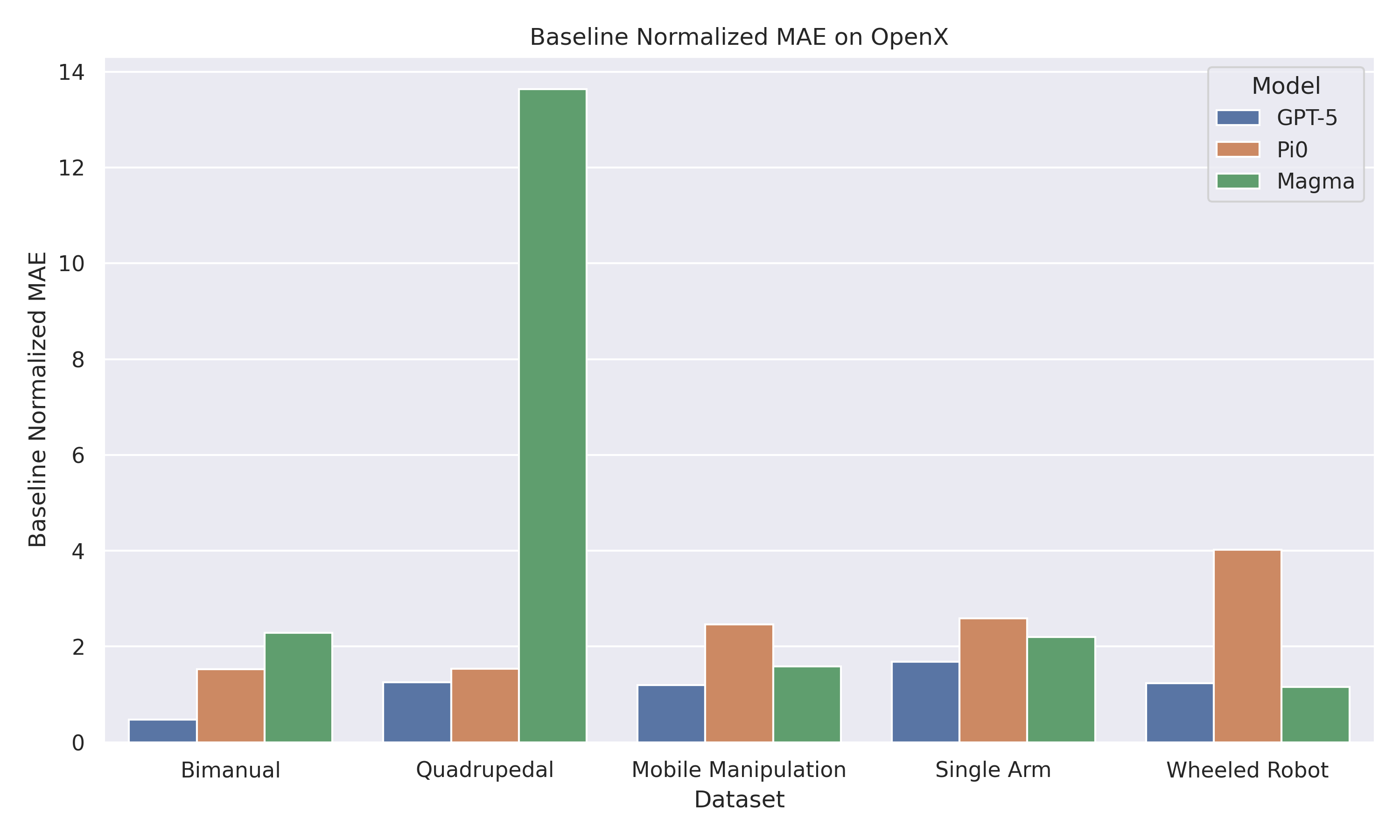}
\caption{Baseline normalized AMAE comparison across action-prediction tasks. GPT-5 yields the lowest aggregated action-prediction error across Open-X, while Magma and $\pi_{0}$ show substantially higher errors.}
\label{fig:openx_bnamae}
\end{figure*}

\begin{figure*}[]
\centering
\includegraphics[width=0.98\linewidth]{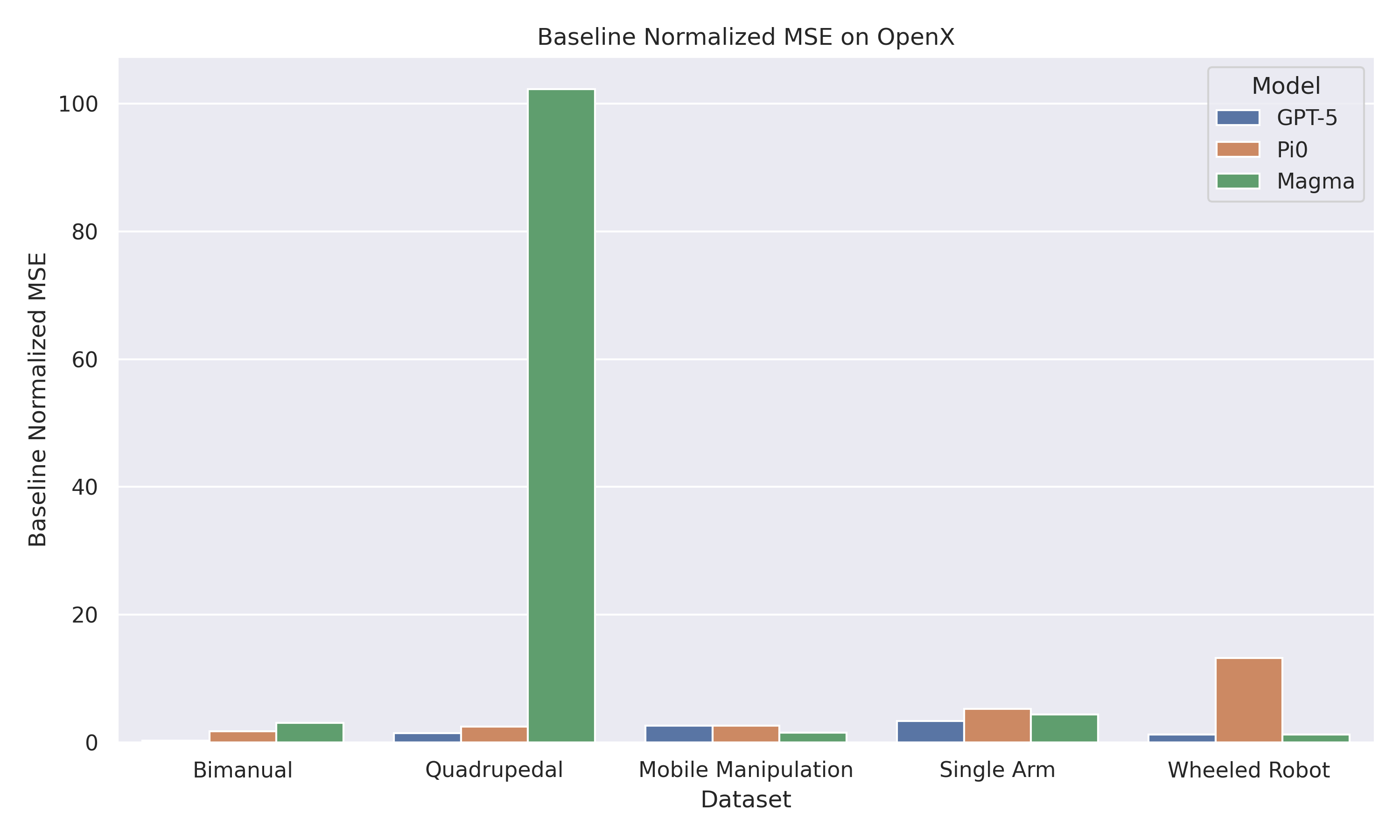}
\caption{Baseline normalized AMSE (NAMSE) comparison across Open-X morphologies, showing persistent challenges in high-DoF continuous control.}
\label{fig:openx_bnamse}
\end{figure*}

Fig.~\ref{fig:openx_bnamae} and Fig.\ref{fig:openx_bnamse} report baseline normalized AMSE (BNAMSE) and baseline normalized AMAE across Open-X morphologies. Surprisingly, GPT-5 achieves the lowest BNAMAE among the three models. However, it is worth noting that GPT-5 is given the action probabilities as part of its prompt. Magma and $\pi_{0}$ perform considerably worse: both models show elevated normalized errors across all embodiments. Notably, Magma's errors on quadrupedal tasks are exceptionally high - substantially above both GPT-5 and $\pi_0$. This is likely due to the fact that the OpenX Quadrupedal dataset is evaluated as a vision language dataset as Magma does not support 12-dimensional action outputs. Since action space transformation methodology for the 12-dimensional OpenX Quadrupedal is not specified in Yang et.al, we prompt Magma to output 12 action values as natural language tokens. 

\subsection{Invalid Outputs and Failure Taxonomy}

Qualitative audits align with the quantitative trends above. On VL tasks, $\pi_0$ degenerates into repeated token fragments (e.g., long “increa increa increa” runs), suggesting a corrupted end-of-sequence behavior acquired during action training. Magma frequently triggers spatial output modes, producing coordinates strings where short textual answers are required. This is likely due to Magma confusing the intended output format with set-of-marks or trace-of-marks as it has been trained to predict next frame set-of-mark action grounding for UI navigation tasks and trace-of-mark action planning for instructional videos \parencite{yang2025magma}. This is not surprising, as some images in the vision language datasets can resemble the images in the instructional videos on which Magma has been trained. the GPT-5 generally preserves validity but can drift toward verbosity, returning plans or explanations when a compact label or number is needed. These failure modes explain the EMR and macro-F1 patterns observed in Figs.~\ref{fig:emr}–\ref{fig:f1}. While $\pi_0$'s failures are repeat-token gibberish and indicates a loss of langauge generation capability, Magma's and GPT-5's failures can be broadly categorized as failure to follow intended output format instructions and can benefit from instruction finetuning. 


Across models, stronger prompts--e.g. providing explicit formats and action-space statistic--improve format compliance but do not erase capability gaps. For example, we qualitatively observed that prompt engineering reduced invalid outputs for GPT-5, yet $\pi_0$ remained at $>90\%$ invalid on VL tasks and Magma still produced coordinates in vision-language tasks—consistent with the low accuracies of RoboVQA, SQA3D and ODINW in Fig.~\ref{fig:emr}.

\section{Conclusion}

MultiNet v1.0 reveals a stark and actionable picture of the current landscape of vision–language–action models: despite rapid progress, today’s systems fail to generalize in any meaningful cross-domain sense. Our evaluation exposes catastrophic cross-domain failure as a fundamental limitation—no model we tested maintains competence across modalities, and in several cases performance collapses to near-zero. Pi0, in particular, drops to 0\% accuracy on basic vision–language tasks, while GPT-5, though stronger, remains far from achieving the reliability required for successful action task execution.

These findings also show that domain-specific fine-tuning corrupts core competencies. Pi0’s repetitive “increa” token emission illustrates how action-oriented adaptation can overwrite language-generation pathways, suggesting that catastrophic forgetting is not an incidental failure mode but a structural consequence of current VLA training pipelines.

We further observe pervasive output-modality misalignment: Magma, despite being framed as a generalist agent, routinely outputs spatial coordinate vectors when asked to answer language questions. This systematic failure indicates that the model does not maintain a coherent mapping between task inputs and expected output formats, underscoring a deeper architectural mismatch rather than simple prompt misuse.

Across all models, prompt engineering provides only marginal benefit, yielding improvements that fail to overcome underlying architectural incompatibilities. This confirms that the barriers to generality stem from model structure and training dynamics, not superficial prompting choices.

Together, these results highlight an urgent need for architectural innovation. Current VLA training paradigms produce overspecialized systems with rooted domain-specific biases that prevent unified reasoning and action across tasks. Moving forward, progress toward true generalist multimodal intelligence will require rethinking model modularity, representation sharing, and progressive training strategies. MultiNet v1.0 provides a foundation for diagnosing these failures and charting a path toward more robust and genuinely cross-domain capable VLA systems.

\clearpage
\printbibliography
\newpage
\appendix

\section{Appendix / supplemental material}
list\ref{lst:bfcl_multiturn} is an example multi-turn conversation that's part of the prompt for the BFCL benchmark.
\lstdefinelanguage{json}{
    basicstyle=\ttfamily\small,
    numbers=left,
    numberstyle=\scriptsize,
    stepnumber=1,
    showstringspaces=false,
    breaklines=true,
    literate=
     *{0}{{{\color{black}0}}}{1}
      {1}{{{\color{black}1}}}{1}
      {2}{{{\color{black}2}}}{1}
      {3}{{{\color{black}3}}}{1}
      {4}{{{\color{black}4}}}{1}
      {5}{{{\color{black}5}}}{1}
      {6}{{{\color{black}6}}}{1}
      {7}{{{\color{black}7}}}{1}
      {8}{{{\color{black}8}}}{1}
      {9}{{{\color{black}9}}}{1}
      {:}{{{\color{black}:{}}}}{1}
      {,}{{{\color{black},{}}}}{1}
      {"}{{{\color{blue}"}}}{1}
      {\{}{{{\color{red}\{}}}{1}
      {\}}{{{\color{red}\}}}}{1}
}

\begin{lstlisting}[language=json, basicstyle=\ttfamily\small, breaklines=true, frame=single, caption={Example multi-turn BFCL prompt used in our evaluation.},label={lst:bfcl_multiturn}]
{
  "conversation": [
    {
      "role": "system",
      "content": "You are an AI assistant that can call functions to complete tasks. You will be presented with conversation histories where each turn may require function calls.

For each turn, analyze the conversation history, which may include previous assistant responses in addition to user prompts, and respond with the correct function to call.
Format each function call as: function_name(param1=value1, param2=value2, ...)
Use only the exact function names available in the provided set of functions and append appropriate parameters.
Output only the function calls, no explanations or additional text."
    },
    {
      "role": "user",
      "content": "Initial Environment Configuration:
{
  "conversation_id": "multi_turn_001",
  "turns": [
    [
      "I need to book a flight from San Francisco to New York for next Monday."
    ],
    [
      "Great! Now book a hotel near JFK airport for the same dates."
    ],
    [
      "Can you also rent a car for the duration of the trip?"
    ]
  ],
  "ground_truth_functions": [
    [
      {
        "function": "book_flight",
        "arguments": {
          "origin": "SFO",
          "destination": "JFK",
          "date": "2024-03-18"
        }
      }
    ],
    [
      {
        "function": "book_hotel",
        "arguments": {
          "location": "near JFK airport",
          "check_in": "2024-03-18",
          "check_out": "2024-03-22"
        }
      }
    ],
    [
      {
        "function": "rent_car",
        "arguments": {
          "pickup_location": "JFK airport",
          "pickup_date": "2024-03-18",
          "return_date": "2024-03-22"
        }
      }
    ]
  ],
  "initial_config": {
    "available_apis": ["booking_service", "travel_assistant"]
  },
  "num_turns": 3
}
\end{lstlisting}

\end{document}